\documentclass{article} % For LaTeX2e
\usepackage{arxiv}
\renewcommand{\undertitle}{}
\renewcommand{\headeright}{}
\renewcommand{\shorttitle}{}
\renewcommand{\headrulewidth}{0pt}
\usepackage{natbib}
\usepackage[T1]{fontenc}

\usepackage{amsmath,amsfonts,bm}

\def\eqref#1{equation~\ref{#1}}
\def\1{\bm{1}}

\DeclareMathAlphabet{\mathsfit}{\encodingdefault}{\sfdefault}{m}{sl}
\SetMathAlphabet{\mathsfit}{bold}{\encodingdefault}{\sfdefault}{bx}{n}

\usepackage{hyperref}
\usepackage{url}
\usepackage{graphicx}
\usepackage{booktabs}
\usepackage{amsmath}
\usepackage{amssymb}
\usepackage{multirow}
\usepackage{xcolor}
\usepackage[capitalise,nameinlink]{cleveref}

\newcommand{\pci}[1]{{\color{gray}\tiny[#1]}}
\title{Beyond the Embedding Bottleneck: Adaptive Retrieval-Augmented 3D CT Report Generation}

\author{
\textbf{Renjie Liang\textsuperscript{1}, Yiling Ma\textsuperscript{2}, Yang Xing\textsuperscript{1}, Zhengkang Fan\textsuperscript{1}, Jinqian Pan\textsuperscript{1},} \\
\textbf{Chengkun Sun\textsuperscript{1}, Shawn Li\textsuperscript{3}, Kuang Gong\textsuperscript{1}, Jie Xu\textsuperscript{1}} \\[4pt]
\small\textsuperscript{1}University of Florida, Gainesville, FL, USA \quad
\textsuperscript{2}Yale University, New Haven, CT, USA \\
\small\textsuperscript{3}University of Southern California, Los Angeles, CA, USA
}
\date{}

\begin{document}

\maketitle

% =====================================================================
% ABSTRACT
% =====================================================================
\begin{abstract}
Automated radiology report generation from 3D CT volumes often suffers from incomplete abnormality coverage. We provide empirical evidence that this limitation stems from a representational bottleneck: contrastive 3D CT embeddings encode discriminative abnormality signals, yet exhibit dimensional concentration, with CT-CLIP concentrating 90\% of its variance in only 2 of 512 principal components. Corroborating this, scaling the language model yields no measurable improvement, suggesting that the bottleneck lies in the visual representation rather than the generator. This bottleneck limits both generation and retrieval. Image-based similarity often fails to isolate fine-grained abnormalities.
We propose \textbf{AdaRAG-CT}, an adaptive augmentation framework that compensates for this visual bottleneck by introducing supplementary textual information through controlled retrieval and selectively integrating it during generation. On the CT-RATE benchmark, AdaRAG-CT reaches a Clinical F1 of 0.480. Ablations show that relevant retrieved context improves clinical efficacy, while adaptive triggering balances clinical and language quality. Code is available at \url{https://github.com/renjie-liang/Adaptive-RAG-for-3DCT-Report-Generation}.
\end{abstract}

% =====================================================================
% 1. INTRODUCTION
% =====================================================================
\section{Introduction}
\label{sec:intro}

Three-dimensional computed tomography (CT) is central to thoracic diagnosis, yet translating CT volumes into structured radiology reports remains labor-intensive. Recent work has explored vision-language models for this task. CT-CHAT~\citep{Hamamci_2026} combines a contrastive 3D vision encoder with a large language model. CT-Agent~\citep{mao2025ctagentmultimodalllmagent3d} incorporates anatomy-aware retrieval and chain-of-thought reasoning. BTB3D~\citep{NEURIPS2025_c4861cc8} replaces contrastive encoders with reconstruction-based tokenization. The best reported Clinical F1 on CT-RATE remains below clinically acceptable levels, indicating that many abnormalities remain unreported.

To understand this limitation, we analyze the representation space of contrastive 3D CT embeddings. Linear probes~\citep{alain2018understandingintermediatelayersusing} achieve AUC scores between 0.59 and 0.97 across 18 abnormalities, confirming that discriminative signals are present. However, principal component analysis shows that most variance is concentrated in only a few dimensions~\citep{jing2022understanding,aghajanyan-etal-2021-intrinsic}, resulting in extremely low effective dimensionality. Organ-level encoders exhibit similar behavior.
These embeddings can therefore distinguish coarse pathological states, but they lack the capacity to represent fine-grained report semantics. Scaling the language model from 8B to 70B parameters yields no measurable improvement~\citep{10655378,NEURIPS2025_c4861cc8}, indicating that the bottleneck lies in the visual representation rather than the decoder.

When the visual channel cannot carry sufficient information, additional information must be introduced through another modality. Retrieval-augmented generation~\citep{NEURIPS2020_6b493230,gao2024retrievalaugmentedgenerationlargelanguage,fink2025retrieval} offers a mechanism for such supplementation: retrieving reference sentences from similar cases and injecting them as context supplies the generator with details not encoded in the visual embedding. However, the reliability of this textual channel depends on the same representation that constrains generation. If the embedding lacks capacity to separate fine-grained abnormalities, image-based retrieval cannot reliably isolate relevant cases. Image-based retrieval has limited abnormality-level precision, while overly frequent context injection can degrade generation.

An empirical analysis across four 3D contrastive encoders confirms this link: abnormality-level retrieval precision remains low, particularly for organs with diverse abnormalities. As a result, image-based retrieval struggles to isolate relevant abnormalities when several co-occur within the same organ. Rather than replacing the encoder, as in concurrent reconstruction-based approaches~\citep{NEURIPS2025_c4861cc8}, we retain existing contrastive models and design a controlled augmentation strategy that maximizes the utility of the textual channel while mitigating the misalignment inherited from the visual representation.

Our framework, AdaRAG-CT, introduces supplementary textual context through controlled retrieval and integrates it selectively during report generation. At the retrieval stage, we explore strategies that improve abnormality-level relevance beyond naive image-based retrieval. At the decoding stage, inspired by Self-RAG~\citep{asai2024selfrag}, a learned \texttt{[RAG]} token allows the model to decide when to retrieve and condition on the returned context, trained via an oracle-mixed regime that exposes the model to both ground-truth and imperfect retrievals.

This work makes three contributions. First, we diagnose a representational bottleneck in 3D medical contrastive embeddings: a PCA analysis across four encoders reveals effective dimensionalities of only a few dimensions, far below natural-image CLIP, providing empirical evidence that the visual channel alone is insufficient in our experiments. Second, we propose AdaRAG-CT, an adaptive augmentation framework that compensates for this bottleneck by introducing and controlling supplementary textual information: retrieval strategies improve abnormality-level relevance, while a learned adaptive mechanism enables the generator to selectively utilize retrieved context. Third, with adaptive retrieval augmentation, our model improves clinical efficacy on CT-RATE over the base model. Ablations show that retrieved context provides additional clinical evidence, while adaptive triggering controls when this context is used.

% =====================================================================
% 2. RELATED WORK
% =====================================================================
\section{Related Work}
\label{sec:related}

\paragraph{3D Medical Report Generation.}
CT2Rep~\citep{hamamci2024ct2rep} introduced the first dedicated 3D CT report-generation framework using an autoregressive transformer with relational memory. Subsequent work paired 3D CT encoders with LLMs (CT-CHAT~\citep{Hamamci_2026}, Med3DVLM~\citep{11145341}), added anatomy-specific LoRA and exemplar retrieval (CT-Agent~\citep{mao2025ctagentmultimodalllmagent3d}), or explored multi-modal fusion~\citep{luo-etal-2025-vividmed,lee2024readlikeradiologistefficient} and memory-driven evidence aggregation~\citep{Chen_2020,10203079}. Several works directly recognise the visual representation as a bottleneck and propose alternative tokenization or aggregation schemes (BTB3D~\citep{NEURIPS2025_c4861cc8}, SAMF~\citep{hosseini2025slices}), or pursue broader 3D understanding via multi-task pre-training (COLIPRI~\citep{wald2026comprehensivelanguageimagepretraining3d}, M3D~\citep{bai2024m3dadvancing3dmedical}, RadFM~\citep{Wu_2025}). Related renal CT work studies lesion characterization as a precursor to clinically grounded report generation~\citep{liang2026kidney,liang2025renal}. Our work complements this line by providing a quantitative diagnosis of \emph{why} existing contrastive embeddings limit retrieval quality, namely that their effective dimensionality is strongly concentrated and poorly aligned with fine-grained abnormality semantics, and proposing an adaptive RAG framework that compensates through a textual channel rather than modifying the encoder.

\paragraph{Retrieval-Augmented Generation in Medical Imaging.}
RAG~\citep{gao2024retrievalaugmentedgenerationlargelanguage} grounds generative models in retrieved evidence, and adaptive variants decide when to retrieve: FLARE~\citep{jiang-etal-2023-active} generates a tentative continuation, detects low-confidence tokens, backtracks, and re-generates with retrieved context. Self-RAG~\citep{asai2024selfrag} trains reflection tokens to critique and select among retrieval-augmented candidates. In medical imaging, retrieval-based approaches have been explored for 2D chest X-ray report generation~\citep{fink2025retrieval,https://doi.org/10.48550/arxiv.2411.16523}, typically retrieving full reports or report sections as templates or auxiliary inputs, and generally assume that retrieval quality is sufficient. Our analysis shows that this assumption breaks down for 3D CT. Unlike FLARE, which triggers retrieval from low-confidence tokens, and Self-RAG, which critiques retrieval-augmented candidates, our model learns to emit a dedicated \texttt{[RAG]} token that decides when to retrieve.

\paragraph{Contrastive Representation Quality.}
Contrastive image--text learning~\citep{pmlr-v139-radford21a} is known to exhibit failure modes in its representation spaces: dimensional collapse, where embeddings occupy a low-dimensional subspace~\citep{jing2022understanding}; modality gap, where image and text embeddings cluster separately~\citep{NEURIPS2022_702f4db7}; and feature suppression, where task-irrelevant variance dominates learned features~\citep{10.1007/978-3-031-73010-8_3}. In the medical domain, CT-CLIP~\citep{Hamamci_2026} and ViSD-Boost~\citep{https://doi.org/10.48550/arxiv.2508.03742} have trained 3D CT encoders with contrastive objectives on CT-RATE, but the retrieval-level consequences of their representation geometry have received little attention. Our analysis complements these works by showing that the dominant variance directions in 3D medical contrastive embeddings fail to encode abnormality semantics, an effect we term \emph{variance-semantic misalignment}, and that this property limits downstream RAG pipelines.

\begin{table}[!htb]
\vspace{-10pt}
\caption{Effective dimensionality and train/validation stability of 3D CT embeddings. Gray brackets denote 95\% intervals. dim$_{90}$/dim$_{95}$ count components explaining 90\%/95\% of variance; PR is the participation ratio.}
\label{tab:pca}
\centering
\scriptsize
\setlength{\tabcolsep}{3.5pt}
\renewcommand{\arraystretch}{0.92}
\resizebox{\textwidth}{!}{%
\begin{tabular}{@{}lllcccc@{}}
\toprule
Model & Embedding & Split & Total dim & dim$_{90}$ & dim$_{95}$ & PR \\
\midrule
CLIP ViT-B/32 (ImageNet) & image emb. & reference & 512 & 243.0 & 322.0 & 64.40 \\
\midrule
\multirow{4}{*}{CT-CLIP (Zero-shot)} & avg pool & train & 512 & 2.0 \pci{2.0,2.0} & 4.0 \pci{4.0,4.0} & 1.43 \pci{1.41,1.45} \\
 & avg pool & valid & 512 & 2.0 \pci{2.0,2.0} & 4.0 \pci{4.0,4.0} & 1.43 \pci{1.41,1.45} \\
 & max pool & train & 512 & 109.8 \pci{108.0,112.0} & 181.3 \pci{180.0,183.0} & 6.70 \pci{6.47,6.93} \\
 & max pool & valid & 512 & 103.6 \pci{102.0,105.0} & 171.1 \pci{169.0,173.0} & 6.65 \pci{6.38,6.86} \\
\midrule
\multirow{4}{*}{CT-CLIP (VocabFine)} & avg pool & train & 512 & 1.0 \pci{1.0,1.0} & 2.0 \pci{2.0,2.0} & 1.16 \pci{1.15,1.17} \\
 & avg pool & valid & 512 & 1.0 \pci{1.0,1.0} & 2.0 \pci{2.0,2.0} & 1.17 \pci{1.16,1.18} \\
 & max pool & train & 512 & 2.5 \pci{2.0,3.0} & 40.7 \pci{36.0,46.0} & 1.26 \pci{1.25,1.27} \\
 & max pool & valid & 512 & 2.8 \pci{2.0,3.0} & 40.4 \pci{35.0,45.0} & 1.26 \pci{1.25,1.28} \\
\midrule
\multirow{4}{*}{ViSD-Boost} & lung & train & 256 & 9.3 \pci{9.0,10.0} & 11.6 \pci{11.0,12.0} & 6.30 \pci{6.12,6.45} \\
 & lung & valid & 256 & 9.1 \pci{9.0,10.0} & 11.6 \pci{11.0,12.0} & 6.26 \pci{6.09,6.45} \\
 & heart & train & 256 & 5.0 \pci{5.0,5.0} & 6.0 \pci{6.0,6.0} & 3.63 \pci{3.53,3.73} \\
 & heart & valid & 256 & 5.0 \pci{5.0,5.0} & 6.0 \pci{6.0,6.0} & 3.62 \pci{3.52,3.72} \\
\midrule
\multirow{4}{*}{FVLM} & lung & train & 256 & 6.0 \pci{6.0,6.0} & 8.0 \pci{8.0,8.0} & 4.94 \pci{4.84,5.03} \\
 & lung & valid & 256 & 6.0 \pci{6.0,6.0} & 7.6 \pci{7.0,8.0} & 5.04 \pci{4.93,5.14} \\
 & heart & train & 256 & 4.0 \pci{4.0,4.0} & 6.0 \pci{6.0,6.0} & 2.59 \pci{2.53,2.65} \\
 & heart & valid & 256 & 4.0 \pci{4.0,4.0} & 6.0 \pci{6.0,6.0} & 2.56 \pci{2.50,2.63} \\
\bottomrule
\end{tabular}
}
\vspace{-10pt}
\end{table}

% =====================================================================
% 3. REPRESENTATIONAL BOTTLENECK ANALYSIS
% =====================================================================
\section{Representational Bottleneck in 3D CT Embeddings}
\label{sec:analysis}

We first diagnose the representational bottleneck of 3D CT contrastive embeddings and its consequences for retrieval. The analysis uses the 46,407 training and 2,987 validation CT-RATE volumes~\citep{Hamamci_2026}, with paired reports and 18 binary abnormality labels, and three families of pre-trained 3D contrastive encoders: CT-CLIP~\citep{Hamamci_2026} and its fine-tuned variant CT-CLIP (VocabFine), FVLM~\citep{shui2025largescalefinegrainedvisionlanguagepretraining}, and ViSD-Boost~\citep{https://doi.org/10.48550/arxiv.2508.03742}.

\begin{table}[!htb]
\vspace{-10pt}
\caption{Linear probe AUC on frozen contrastive embeddings.\textsuperscript{*}}
\label{tab:linear_probe}
\centering
\scriptsize
\begin{tabular}{@{}lcccccc@{}}
\toprule
& \multicolumn{2}{c}{CT-CLIP (Zero-shot)} & \multicolumn{2}{c}{CT-CLIP (VocabFine)} & & \\
\cmidrule(lr){2-3} \cmidrule(lr){4-5}
Abnormality & avg & max & avg & max & FVLM & ViSD-Boost \\
\midrule
\multicolumn{7}{@{}l}{\textit{Lung (11)}} \\
Lung nodule        & 0.619 & 0.582 & 0.588 & 0.516 & 0.664 & 0.713 \\
Mosaic attenuation & 0.794 & 0.710 & 0.711 & 0.578 & 0.887 & 0.892 \\
Peribronchial thick. & 0.717 & 0.631 & 0.730 & 0.645 & 0.777 & 0.814 \\
Consolidation      & 0.732 & 0.659 & 0.630 & 0.543 & 0.859 & 0.920 \\
Bronchiectasis     & 0.679 & 0.605 & 0.624 & 0.564 & 0.752 & 0.790 \\
Interlob.\ septal thick. & 0.762 & 0.700 & 0.725 & 0.583 & 0.863 & 0.874 \\
Emphysema          & 0.712 & 0.629 & 0.695 & 0.641 & 0.792 & 0.799 \\
Atelectasis        & 0.680 & 0.627 & 0.667 & 0.561 & 0.738 & 0.797 \\
Lung opacity       & 0.653 & 0.610 & 0.634 & 0.551 & 0.805 & 0.871 \\
Pulm.\ fibrotic seq. & 0.621 & 0.582 & 0.614 & 0.542 & 0.688 & 0.744 \\
Pleural effusion   & 0.911 & 0.849 & 0.850 & 0.718 & 0.956 & 0.966 \\
\midrule
\multicolumn{7}{@{}l}{\textit{Heart (2)}} \\
Cardiomegaly          & 0.891 & 0.794 & 0.836 & 0.666 & 0.943 & 0.952 \\
Pericardial effusion  & 0.797 & 0.687 & 0.764 & 0.685 & 0.851 & 0.910 \\
\midrule
\multicolumn{7}{@{}l}{\textit{Aorta (2)}} \\
Arterial wall calc.   & 0.818 & 0.734 & 0.808 & 0.676 & 0.898 & 0.913 \\
Coronary artery calc. & 0.812 & 0.737 & 0.797 & 0.664 & 0.879 & 0.903 \\
\midrule
\multicolumn{7}{@{}l}{\textit{Esophagus (1)}} \\
Hiatal hernia    & 0.701 & 0.650 & 0.665 & 0.573 & 0.758 & 0.821 \\
\midrule
\multicolumn{7}{@{}l}{\textit{Other (2)}} \\
Lymphadenopathy  & 0.688 & 0.621 & 0.648 & 0.586 & ---   & ---   \\
Medical material & 0.726 & 0.642 & 0.688 & 0.590 & ---   & ---   \\
\midrule
\textbf{Macro avg} & \textbf{0.739} & \textbf{0.669} & \textbf{0.704} & \textbf{0.604} & \textbf{0.823} & \textbf{0.853} \\
\bottomrule
\end{tabular}
\begin{flushleft}
\textsuperscript{*} ``---'': abnormality outside the organ encoder's scope.
\end{flushleft}
\vspace{-10pt}
\end{table}

\subsection{Encoded but Narrow}
\label{sec:analysis-encoded}

Following \citet{alain2018understandingintermediatelayersusing}, we train linear probes (logistic regression) on frozen representations to test whether abnormality information is linearly accessible. Across the three encoders, probe AUC ranges from 0.59 to 0.97 for average-pooled embeddings across 18 abnormalities (\cref{tab:linear_probe}), confirming that the contrastive objective successfully encodes diagnostic signals.
However, high AUC does not imply rich information content. A principal component analysis~\citep{jing2022understanding,aghajanyan-etal-2021-intrinsic} of the embedding spaces reveals that the effective dimensionality is remarkably low (\cref{tab:pca}). For each train/validation split, we report means and 95\% intervals over 200 bootstrap resamples of 2{,}987 volumes. CT-CLIP with average pooling concentrates 90\% of its variance in only 2 principal components (dim$_{90}$\,=\,2, PR\,=\,1.4). Although max pooling inflates dim$_{90}$ to about 110, the participation ratio remains only 6.7, indicating that most of this spread is long-tail noise rather than signal. Consistent with this, zero-shot CT-CLIP's macro AUC is lower for max pooling than average pooling (0.669 vs.\ 0.739; see \cref{tab:linear_probe}). Organ-level embeddings from ViSD-Boost and FVLM fare similarly (dim$_{90}$ is approximately 4--9 out of 256). A probe on the top 2 principal components of zero-shot CT-CLIP reaches a mean AUC of only 0.552, against 0.746 on the lower half. An embedding can achieve high binary AUC (e.g., ``pleural effusion present vs.\ absent'') using just one or two directions, yet may not reliably preserve the fine-grained semantic detail required for report generation, such as laterality, severity, and co-occurring abnormalities. This distinction between discriminability (high AUC) and information richness (low effective dimensionality) is central to our argument: the bottleneck is not that the LLM fails to extract available information, but that \emph{the visual channel is too limited} to begin with.
A natural-image baseline confirms the severity of this concentration: CLIP ViT-B/32 on ImageNet achieves dim$_{90}$\,=\,243 and PR\,=\,64.4 on the same 512-dimensional space, about 45$\times$ higher in PR than CT-CLIP (\cref{tab:pca}). Two factors likely explain the gap. First, natural images span visually diverse categories, encouraging the encoder to use many dimensions. Chest CT volumes are structurally homogeneous, with pathological differences confined to subtle local variations, so a contrastive objective can separate most cases with very few directions. Second, existing 3D CT encoders are trained with a global contrastive loss that pairs an entire volume with an entire report, a granularity too coarse to incentivize fine-grained, abnormality-level representations.
A scaling experiment corroborates this diagnosis: replacing the 8B LLM with a 70B variant yields Clinical F1 of 0.405 versus 0.455, showing no improvement despite an 8.75$\times$ increase in generator capacity. If the bottleneck were in the decoder, a larger model should improve performance. The absence of any gain confirms that the visual channel is the limiting factor. This observation is consistent with prior work: CT-CHAT reports near-identical metrics at 8B and 70B scales~\citep{Hamamci_2026}, and BTB3D similarly concludes that improved visual tokenization, rather than larger language backbones, is essential for 3D medical VLMs~\citep{NEURIPS2025_c4861cc8}.

\begin{table}[!htb]
\vspace{-10pt}
\caption{Organ-level retrieval precision (Jaccard@10) with ViSD-Boost image embeddings and its text encoder on organ paragraphs. Upper Bound ranks by ground-truth label overlap.}
\label{tab:retrieval}
\centering
\scriptsize
\begin{tabular}{@{}lcccc@{}}
\toprule
Organ & Img2Img & Img2Txt & Txt2Txt & Upper Bound \\
\midrule
Lung & 0.351 & 0.313 & 0.563 & 0.992 \\
Heart & 0.825 & 0.638 & 0.925 & 1.000 \\
Esophagus & 0.796 & 0.617 & 0.935 & 1.000 \\
Aorta & 0.741 & 0.617 & 0.666 & 1.000 \\
\bottomrule
\end{tabular}
\vspace{-10pt}
\end{table}

\subsection{Retrieval Under the Bottleneck}
\label{sec:analysis-retrieval}

The dimensional concentration identified above has important consequences for retrieval. When most variance concentrates in a few dimensions, cosine similarity is dominated by these directions, collapsing fine-grained abnormality distinctions. We measure organ-level Jaccard@10, the intersection over union of abnormality label sets between the query and each of the top-10 cosine-nearest neighbours averaged over queries, and compare image-to-image, image-to-text (cross-modal), and text-to-text retrieval against an upper bound constructed from ground-truth abnormality label overlap (\cref{tab:retrieval}).
Lung retrieval is weakest (Jaccard@10 = 0.351): with 11 distinct abnormalities, the Jaccard denominator grows quickly, requiring precise alignment across a combinatorially large label space. The upper bound of 0.992 for lung confirms that the database contains relevant matches. The bottleneck lies in the representation, not the coverage. By contrast, heart (2 abnormalities) and esophagus (1 abnormality) achieve much higher overlap simply because fewer labels need to agree. The aorta is an exception (Img2Img 0.741 $>$ Txt2Txt 0.666): its pathological vocabulary (arterial and coronary calcification) overlaps substantially with the heart section, reducing text-to-text discriminability, whereas the organ-specific image embeddings remain well-separated by construction.

A clear asymmetry emerges across modalities. Text-to-text cosine similarity outperforms image-to-image retrieval for lung (0.563 vs.\ 0.351), heart (0.925 vs.\ 0.825), and esophagus (0.935 vs.\ 0.796), indicating that the text encoder preserves abnormality semantics more effectively than the image encoder. Cross-modal retrieval (Img2Txt) is worse than image-to-image for all organs, indicating that direct cross-modal matching is unreliable in this setting. The gap between Txt2Txt and Img2Img suggests that the bottleneck originates in the image representation, not in cosine similarity per se. Conventional medical RAG systems query a text database with the image embedding (Img2Txt), which is directly constrained by this bottleneck. Text-to-text retrieval offers a way to reduce this dependence, suggesting that retrieval strategies using partial text may be more reliable than purely image-based approaches.

% =====================================================================
% 4. METHOD
% =====================================================================
\section{Method: AdaRAG-CT}
\label{sec:method}

\begin{figure}[!t]
\vspace{-10pt}
  \centering
  \includegraphics[width=0.90\linewidth]{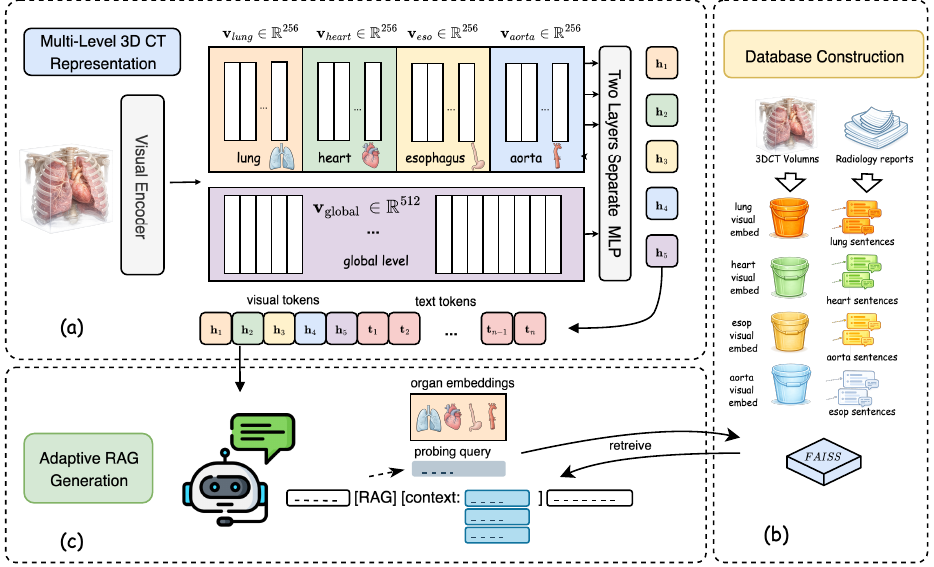}
  \caption{Overview of AdaRAG-CT.
(a) A 3D CT volume is encoded into a global embedding (CT-CLIP) and four organ-specific embeddings (ViSD-Boost), which are projected into the LLM input space as visual tokens. (b) An organ-indexed sentence database is built from training-set reports and indexed with FAISS for efficient retrieval. (c) During report generation, the LLM autonomously emits a \texttt{[RAG]} token when it needs external evidence. Retrieved sentences are then injected into the context before the model continues generating.}
  \label{fig:method}
\vspace{-10pt}
\end{figure}

AdaRAG-CT operates on top of a base vision-language model and augments it with adaptive retrieval-augmented generation.

\subsection{Base Model}
\label{sec:method-base}

We adopt a LLaVA-style architecture~\citep{NEURIPS2023_6dcf277e} where a pre-trained LLM (Llama-3.1-8B-Instruct and Llama-3.3-70B-Instruct~\citep{grattafiori2024llama3herdmodels}) receives visual input through learned projectors. Each 3D CT volume is represented by $K=5$ embeddings: a global max-pooled embedding $\mathbf{v}_{\text{global}} \in \mathbb{R}^{512}$ from CT-CLIP and four organ-specific embeddings $\mathbf{v}_{o} \in \mathbb{R}^{256}$ ($o \in \{\text{lung, heart, esophagus, aorta}\}$) from ViSD-Boost. Each embedding is projected independently as
\begin{equation}
  \mathbf{h}_o = \text{MLP}_o(\mathbf{v}_o), \quad \mathbf{h}_o \in \mathbb{R}^{d_{\text{LLM}}},
  \label{eq:projector}
\end{equation}
where $\text{MLP}_o$ is a 2-layer projector specific to organ $o$. The resulting visual tokens $\{\mathbf{h}_o\}_{o=1}^{K}$ are incorporated into the input token sequence as visual context for the LLM. This base model, without any retrieval augmentation, serves as our primary baseline. This strong base is the result of design experiments on embedding combinations, organ-specific projectors, and two-stage LoRA training.

\subsection{Organ-Indexed Sentence Database}
\label{sec:method-database}

We construct an organ-indexed sentence database from all training reports. Each report is first split into individual sentences. Every sentence is then annotated with its corresponding organ label (lung, heart, esophagus, aorta) and associated abnormalities, yielding approximately 572K sentences in total across both Findings and Impression sections. To represent each sentence, we employ a text encoder based on BiomedVLP-CXR-BERT~\citep{10.1007/978-3-031-20059-5_1}, followed by a learned 256-dimensional projection layer. The text encoder and projection layer are fine-tuned using a contrastive image--text objective, with frozen organ-level image embeddings serving as anchors. All sentence embeddings are stored in per-organ FAISS indices to enable efficient nearest-neighbour retrieval. Organ-level image embeddings from the training set are indexed for coarse candidate retrieval.

\subsection{Adaptive Retrieval Training}
\label{sec:method-training}

Following Self-RAG~\citep{asai2024selfrag}, we introduce a special \texttt{[RAG]} token into the vocabulary. At inference, rather than injecting context at predetermined positions, the model generates \texttt{[RAG]} to request external context. During training, we run the base model on ground-truth reports and compute per-sentence perplexity. Sentences above the 70th percentile ($\mathrm{ppl}\approx1.935$) are selected as retrieval targets, with at most four targets per report. Intuitively, these high-perplexity sentences are the parts that the base generator finds difficult under visual conditioning alone, and therefore mark positions where additional evidence is likely to help. This threshold is used only to construct training supervision.
For each target, we precompute oracle context using the ground-truth target sentence as a Text2Text query, and real context using image-based coarse filtering followed by text re-ranking. Final candidates are selected using Maximal Marginal Relevance (MMR)~\citep{10.1145/290941.291025}:

\begin{equation}
  \text{MMR}(d_i) = \lambda \cdot \text{sim}(d_i) - (1 - \lambda) \cdot \max_{d_j \in \mathcal{S}} \text{BLEU-2}(d_i, d_j),
  \label{eq:mmr}
\end{equation}
where $\text{sim}(d_i)$ is the cosine similarity score and $\mathcal{S}$ is the set of already-selected sentences. Each training report receives oracle context with probability 0.7 and real context with probability 0.3, for both model sizes. This oracle-mixed regime exposes the model to high-quality signals while building robustness to imperfect retrieval. Oracle context is used only during training. Injected context tokens are masked from the generation loss.

\subsection{Adaptive Retrieval Inference}
\label{sec:method-retrieval}

During autoregressive generation, whenever the model emits \texttt{[RAG]}, it continues to generate the next sentence, which is then used as the retrieval query. The retrieved sentences are formatted and injected into the context, the initial generation is rolled back, and the model regenerates the sentence with the retrieved context.
We explore two retrieval strategies for supplying candidate sentences. As shown in \cref{fig:retrieval_comparison}, both yield comparable generation quality, indicating that performance is robust to the choice of retrieval pipeline. Both strategies use MMR (\cref{eq:mmr}) with $K_{\text{fine}} = 3$.

\paragraph{Two-Stage retrieval.}
Given the organ currently being described, we first retrieve the $K_{\text{coarse}} = 20$ most similar training samples using image embedding cosine similarity, collecting all their sentences for the same organ as a candidate pool. This vision-guided stage narrows the search from the full database to patient-relevant candidates. Candidates are then re-ranked and selected via MMR (\cref{eq:mmr}) against the current query sentence.

\paragraph{Text2Text retrieval.}
We retrieve directly from the full per-organ sentence database using text-to-text cosine similarity, with the model's partial generation as query. Because this query already reflects the organ and tentative abnormality being described, Text2Text retrieval is less dependent on the low-dimensional image embedding. It is still conditioned on the generated report context, and therefore complements rather than replaces the visual input.

% =====================================================================
% 5. EXPERIMENTS
% =====================================================================
\section{Experiments}
\label{sec:experiments}

\subsection{Setup}
\label{sec:exp-setup}

\paragraph{Dataset.} All experiments use the official CT-RATE~\citep{Hamamci_2026} training and validation splits. After excluding CT volumes without pre-computed organ embeddings, we use 46,407 of 47,149 training volumes and 2,987 of 3,039 validation volumes. Each volume is paired with a radiology report, which we concatenate its Findings and Impression sections as the generation target.
\paragraph{Metrics.} Following the CT-RATE evaluation protocol~\citep{Hamamci_2026}, we adopt Clinical Efficacy as the primary metric. A RadBERT-based classifier~\citep{doi:10.1148/ryai.210258} extracts 18 binary abnormality labels (e.g., lung nodule, pleural effusion, cardiomegaly) from each generated report. These predictions are compared against the dataset-provided ground-truth annotations to compute per-abnormality Precision, Recall, and F1. We report the support-weighted average as Clinical Precision (Clin-P), Clinical Recall (Clin-R), and Clinical F1 (Clin-F1). We also report BLEU~\citep{papineni-etal-2002-bleu}, ROUGE-L~\citep{lin-2004-rouge}, and METEOR~\citep{banerjee-lavie-2005-meteor} for textual overlap, and LLaMA Score~\citep{Hamamci_2026}, a report quality rating by Llama-3.1-70B-Instruct.

\paragraph{Baselines.} We compare against prior methods on CT-RATE report generation: CT2Rep~\citep{hamamci2024ct2rep}, an encoder-decoder with relational memory; Merlin~\citep{blankemeier2024merlin}, a 7B VLM for 3D CT understanding; CT-CHAT~\citep{Hamamci_2026}, which pairs a CT-ViT encoder with Llama-3.1 (8B/70B) via attention pooling; BTB3D~\citep{NEURIPS2025_c4861cc8}, which uses reconstruction-based visual tokenization; SAMF~\citep{hosseini2025slices}, a 2D--3D feature fusion approach; and CT-Agent~\citep{mao2025ctagentmultimodalllmagent3d}, an agent-based pipeline with chain-of-thought reasoning. All baseline numbers are taken from the original publications. We also report our own base model (no RAG) at 8B and 70B scales, trained in two stages: organ-specific projectors with frozen LLM weights, followed by LoRA fine-tuning.

\subsection{Main Results}
\label{sec:exp-main}

\begin{table}[!htb]
\vspace{-10pt}
\caption{Report generation performance on the CT-RATE validation set. Baseline numbers are from original publications.}
\label{tab:main}
\centering
\scriptsize
\setlength{\tabcolsep}{3pt}
\begin{tabular}{@{}ll c ccc c cccc c cc @{}}
\toprule
& & & \multicolumn{3}{c}{Clinical Efficacy} & & \multicolumn{4}{c}{NLP} & & \\
\cmidrule(lr){4-6} \cmidrule(lr){8-11} 
Method & Params & & F1 & P & R & & B-1 & B-4 & R-L & MET & & LLaMA Score \\
\midrule
CT2Rep~\citep{hamamci2024ct2rep} & --- & & 0.160 & 0.435 & 0.128 & & 0.372 & 0.213 & --- & 0.197 & & --- \\
Merlin~\citep{blankemeier2024merlin} & 7B & & 0.160 & 0.295 & 0.112 & & 0.231 & 0.099 & --- & 0.148 & & --- \\
CT-CHAT~\citep{Hamamci_2026} & 8B & & --- &  --- & --- & & 0.494 & --- & 0.584 & 0.311 & & 7.440 \\
CT-CHAT~\citep{Hamamci_2026} & 70B & & 0.184 & 0.450 & 0.158 & & 0.498 & --- & 0.581 & 0.311 & & 7.429 \\
BTB3D~\citep{NEURIPS2025_c4861cc8} & 8B & & 0.258 & 0.260 & 0.260 & & 0.439 & 0.213 & --- & 0.223 & & --- \\
SAMF~\citep{hosseini2025slices} & 3.8B & & --- & --- & --- & & 0.440 & 0.261 & 0.417 & 0.417 & & 7.165 \\
CT-Agent~\citep{mao2025ctagentmultimodalllmagent3d} & --- & & 0.420 & 0.423 & 0.477 & & 0.502 & 0.231 & 0.490 & 0.425 & & --- \\
\midrule
Ours (base) & 8B & & 0.455 & 0.474 & 0.469 & & 0.463 & 0.205 & 0.315 & 0.206 & & 7.30 \\
\textbf{AdaRAG-CT} & \textbf{8B} & & \textbf{0.480} & \textbf{0.502} & \textbf{0.520} & & \textbf{0.496} & \textbf{0.242} & \textbf{0.354} & \textbf{0.246} & & \textbf{7.75} \\
\midrule
Ours (base) & 70B & & 0.405 & 0.468 & 0.373 & & 0.449 & 0.213 & 0.334 & 0.208 & & 7.10 \\
\textbf{AdaRAG-CT} & \textbf{70B} & & \textbf{0.426} & \textbf{0.483} & \textbf{0.413} & & \textbf{0.497} & \textbf{0.250} & \textbf{0.361} & \textbf{0.232} & & \textbf{7.53} \\
\bottomrule
\end{tabular}
\begin{flushleft}
\scriptsize Note: Additional baseline results under our unified evaluation protocol are provided in the Supplementary Material.
\end{flushleft}
\vspace{-10pt}
\end{table}

\Cref{tab:main} presents the main comparison. Our 8B base model obtains a Clinical F1 of 0.455 without retrieval. Scaling the LLM to 70B does not improve this figure. In fact, the 70B base model scores only 0.405, consistent with the PCA analysis showing severe dimensional concentration in the visual embeddings (\cref{sec:analysis-encoded}). Both models read the same impoverished visual representations, so the larger decoder has no additional signal to exploit. The bottleneck is representational, not generative.
Retrieval augmentation benefits both scales: the 8B model improves from 0.455 to 0.480 (+0.025) and the 70B model from 0.405 to 0.426 (+0.021), confirming that the textual channel supplies information absent from the visual input. The 8B AdaRAG-CT (0.480) surpasses CT-Agent (0.420) and substantially outperforms CT-CHAT 70B (0.184) in Clinical F1, while also improving NLP metrics and LLaMA Score over our 8B base model.

\subsection{Retrieval Pipeline Comparison}
\label{sec:exp-retrieval-pipeline}

\begin{figure}[!b]
\vspace{-10pt}
  \centering
  \includegraphics[width=\linewidth]{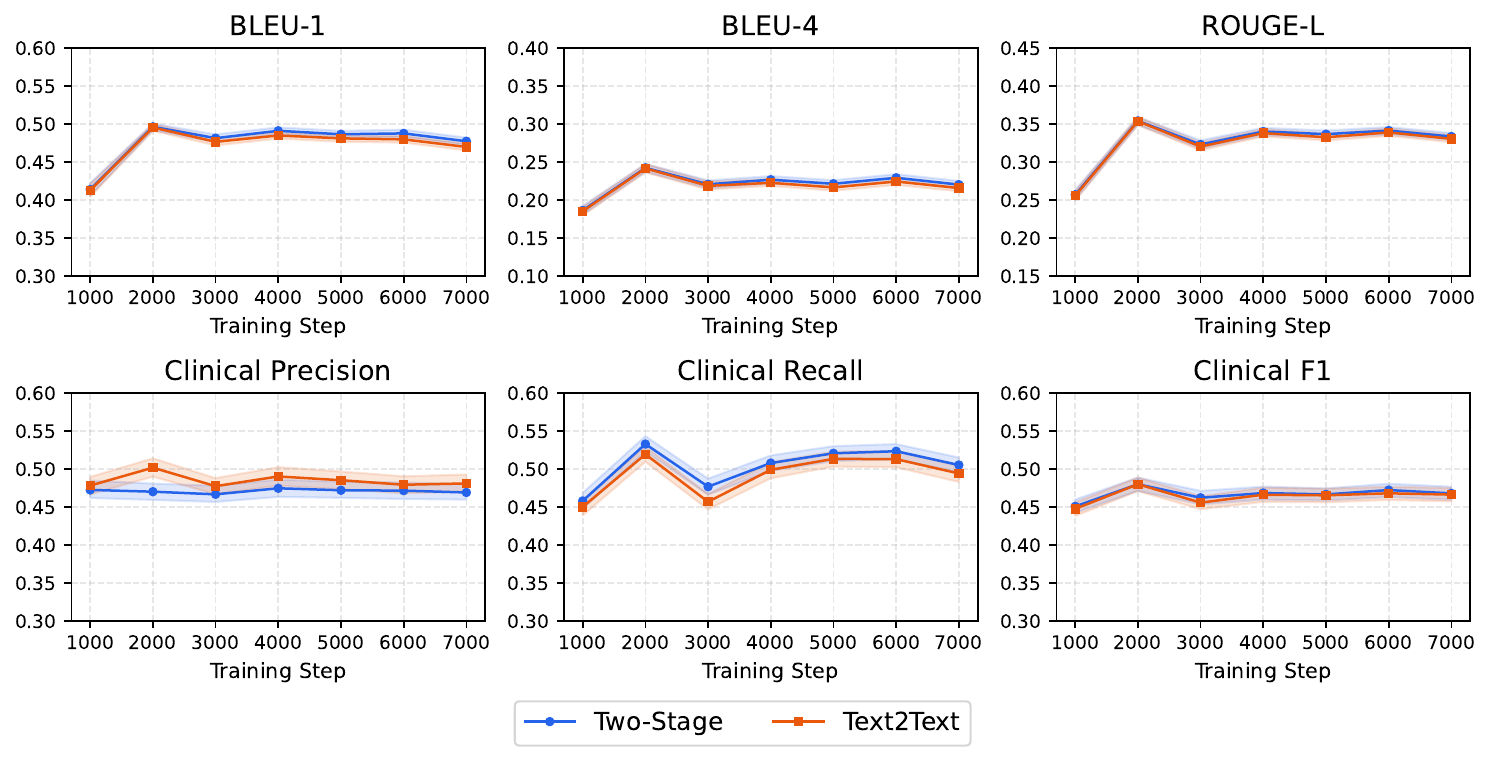}
  \caption{Six evaluation metrics across training steps for Two-Stage and Text2Text retrieval pipelines on the 8B model, with bootstrap 95\% confidence intervals.}
  \label{fig:retrieval_comparison}
\vspace{-10pt}
\end{figure}

\Cref{fig:retrieval_comparison} shows that both pipelines yield stable performance throughout training, with Clinical F1 virtually identical at every step and bootstrap confidence intervals consistently overlapping. A closer look at the precision--recall decomposition reveals a mild trade-off: Two-Stage retrieval, which first narrows candidates via image similarity before text re-ranking, tends to surface a broader set of organ-level abnormalities and thus favours recall. Text2Text retrieval, which matches directly against the model's partial generation, returns more topically precise context and favours precision. These complementary biases cancel out in F1, showing that AdaRAG-CT remains effective across both retrieval pipelines. This also explains why neither pipeline should be interpreted as strictly superior: they expose different useful contexts to the same generator, and the adaptive trigger determines when those contexts enter the report. This robustness is consistent with the bottleneck hypothesis: clinically relevant textual context can compensate for the limited visual signal under either retrieval design.

\subsection{Context Utilization Ablation}
\label{sec:exp-utilization-ablation}

\begin{table}[!htb]
\vspace{-10pt}
\caption{Context utilization ablation (8B model). All variants share the same retrieval pipeline. $\Delta$ is relative to the retrieval-disabled setting.}
\label{tab:rag_ablation}
\centering
\scriptsize
\begin{tabular}{@{}lcccc@{}}
\toprule
Utilization Strategy & Clin-F1 & $\Delta$ & BLEU-4 & ROUGE-L \\
\midrule
Retrieval disabled & 0.462 & --- & 0.228 & 0.349 \\
\midrule
Fixed-interval ($N$=3) & 0.453 & $-$0.009 & 0.192 & 0.307 \\
Fixed-interval ($N$=5) & \textbf{0.494} & \textbf{+0.032} & 0.209 & 0.329 \\
Fixed-interval ($N$=7) & 0.482 & +0.020 & 0.228 & 0.346 \\
Adaptive RAG (ours) & 0.480 & +0.018 & \textbf{0.242} & \textbf{0.354} \\
\bottomrule
\end{tabular}
\vspace{-10pt}
\end{table}

\Cref{tab:rag_ablation} ablates the context injection strategy while holding the retrieval pipeline fixed. The retrieval-disabled setting (0.462) reflects the RAG-trained model's own generation capability without retrieval context at inference, which differs slightly from the base model in \cref{tab:main} (0.455) that was trained without the RAG objective. Fixed-interval injection is sensitive to the choice of $N$: injecting too frequently ($N$=3) hurts performance below baseline, while $N$=5 maximises Clinical F1 (0.494) at the expense of NLP quality, and $N$=7 strikes a middle ground. This sensitivity to $N$ highlights a practical limitation of fixed-interval strategies. The optimal frequency is dataset-dependent and requires separate tuning. Adaptive RAG avoids fixed-interval tuning by triggering retrieval selectively: it achieves competitive Clinical F1 (0.480, +0.018) without any frequency hyperparameter, and consistently outperforms all fixed-interval variants on BLEU-4 and ROUGE-L. On average, the model emits 1.48 \texttt{[RAG]} triggers per report, confirming that retrieval is used sparingly and on demand rather than applied uniformly.

% =====================================================================
% 6. DISCUSSION
% =====================================================================
\section{Discussion}
\label{sec:discussion}

\textbf{Representational bottleneck.} The PCA and scaling results place the bottleneck in the visual representation. Retrieved text therefore acts as a second channel into the generator, one that can carry laterality, severity and co-occurring abnormalities that the embedding does not encode. This differs from the usual role of RAG, which is to supply external world knowledge. Reshaping the encoder directly did not help. Fine-tuning it with a PCA regulariser gave no stable gain in zero-shot classification or in report generation, which suggests that low effective dimensionality reflects missing lesion-level semantics, and a spectrum penalty alone does not supply them.

\textbf{Adaptive triggering.} The retrieval module opens the textual channel and the \texttt{[RAG]} token decides when it is used. Fixed-interval injection depends on its frequency: $N=5$ reaches a higher Clinical F1 than adaptive triggering (0.494 against 0.480) but lower BLEU-4 and ROUGE-L, and $N$ must be tuned per dataset. Oracle-mixed training is what makes the trigger usable. Seeing both oracle and real context teaches the generator to use retrieved evidence where it asks for it and to generate normally elsewhere. The mixing probability $p_{\text{oracle}}$ is itself a hyperparameter whose best value may depend on the dataset. We also expect a 70B generator, with stronger language priors, to produce more confident hallucinations from the same visual input, which case-relevant evidence counters.

\textbf{Retrieved content.} The database is built from the training reports and adds no external medical knowledge. Because its sentences share the generator's training distribution, our results cannot fully separate the value of retrieval from the proximity of the retrieved cases to the training population. Many sentences are templates such as ``No evidence of pleural effusion'', which can fire the trigger without adding case-specific evidence and pull the output toward common phrasing. A median split on overlap with the database is consistent with this. High-overlap reports score higher on BLEU-4 and ROUGE-L (0.246 and 0.363 against 0.238 and 0.345) and lower on Clinical F1 (0.475 against 0.482), so template reuse mainly changes the wording.

\textbf{Scope.} All experiments use CT-RATE, a single-institution set of non-contrast chest CT, and other datasets and anatomical regions remain untested. Clinical F1 scores the presence of 18 labels. It is blind to size, location, laterality and severity, to temporal comparison and to abnormalities outside the 18 labels, so a report that flags a lung nodule without its size scores the same as one that gives it. The evaluation is retrospective, and a reader study is needed before clinical use.

% =====================================================================
% 7. CONCLUSION
% =====================================================================
\section{Conclusion}
\label{sec:conclusion}

Across 3D CT contrastive embeddings, PCA reveals severe dimensional concentration. CT-CLIP concentrates 90\% of its variance in only 2 of 512 principal components. This representational poverty, not the generator's capacity, is the fundamental bottleneck: scaling the LLM from 8B to 70B parameters yields no improvement because both models read the same impoverished visual channel.
AdaRAG-CT addresses this bottleneck by opening a supplementary textual channel through adaptive retrieval. A learned \texttt{[RAG]} token triggers case-relevant context injection during generation, achieving a Clinical F1 of 0.480 on CT-RATE. Ablations show that clinically relevant retrieved context provides the main clinical gain, while adaptive triggering offers a balanced way to control when this context is introduced.
These results suggest that future 3D CT report-generation systems should evaluate not only decoder scale, but also the amount of clinically useful information available to the generator.
Resolving the bottleneck at its source, through abnormality-aware visual pre-training or dense per-token visual representations, remains the longer-term solution. Until richer encoders become available, the textual retrieval channel provides an effective surrogate.

\section*{Acknowledgements}
This work was supported by National Institutes of Health (NIH)/National Cancer Institute (NCI) (R01CA303940).

% =====================================================================
\clearpage
\bibliography{main}
\bibliographystyle{plainnat}

\newpage
\appendix

\begin{center}
{\Large\bfseries
Beyond the Embedding Bottleneck: Adaptive Retrieval-Augmented 3D CT Report Generation
}

\vspace{0.5em}

{\large\bfseries Supplementary Material}
\end{center}

\vspace{1em}

\section{Additional Evidence for the Embedding Bottleneck}
\label{sec:supp-analysis}

\subsection{PCA and linear probe protocol details}
We evaluate representational bottleneck severity using both principal component analysis (PCA) and frozen linear probes. PCA is computed via full SVD on mean-centred embeddings. We report three effective-dimensionality metrics: dim$_{90}$ and dim$_{95}$ (the minimum number of principal components needed to explain 90\% and 95\% of total variance, respectively), and the participation ratio
\begin{equation}
  \mathrm{PR} = \frac{\bigl(\sum_i \sigma_i^2\bigr)^2}{\sum_i \sigma_i^4},
\end{equation}
where $\sigma_i$ are the singular values. PR equals the embedding dimension when all dimensions carry equal variance and approaches 1 under complete collapse.

Linear probes are trained independently for each of the 18 CT-RATE abnormality labels. Each probe is a logistic regression (solver: L-BFGS, $C{=}1.0$, class-balanced weighting, up to 1{,}000 iterations) trained on the training set and evaluated on the held-out validation set. Classification performance is reported as AUC-ROC.

\subsection{Projection test details}
To quantify variance-semantic misalignment, we perform a \emph{projection test}: for each class or abnormality, we train a linear probe on only the top-$k$ or tail-$k$ principal components of the embedding space and measure binary classification AUC. If the dominant variance directions encode class-discriminative information, the top-$k$ probe should outperform the tail-$k$ probe. A reversal (tail $>$ top) indicates that discriminative signals reside in low-variance dimensions that cosine similarity effectively ignores.

\Cref{tab:supp-crossdomain} compares CLIP ViT-B/32 on ImageNet with the zero-shot CT-CLIP embedding used in our pipeline. Both domains exhibit the tail $>$ top pattern, but the severity differs substantially. For CLIP, the top-2 PCs remain informative (AUC 0.82), whereas for CT-CLIP zero-shot the top-2 AUC is only 0.55, near random chance. This supports the claim that 3D medical contrastive embeddings suffer a much stronger form of variance-semantic misalignment than natural-image CLIP.

\begin{table}[!htb]
\vspace{-10pt}
\caption{Cross-domain projection test (mean over all classes or abnormalities). top2: AUC using only the 2 highest-variance PCs. tail$_{1/2}$: AUC using the lower half of all PCs. $\Delta$: tail$_{1/2}${} $-$ top2.}
\label{tab:supp-crossdomain}
\centering
\scriptsize
\begin{tabular}{@{}lccc@{}}
\toprule
Domain / Model & top2 AUC & tail$_{1/2}${} AUC & $\Delta$ \\
\midrule
CLIP ViT-B/32 (ImageNet, 30 classes) & 0.821 & 1.000 & +0.178 \\
CT-CLIP zero-shot, avg pool (CT-RATE, 18 abnormalities) & 0.552 & 0.746 & +0.193 \\
CT-CLIP zero-shot, max pool (CT-RATE, 18 abnormalities) & 0.533 & 0.739 & +0.206 \\
\bottomrule
\end{tabular}
\vspace{-10pt}
\end{table}

\Cref{tab:supp-perfinding} shows the projection test for all 18 CT-RATE abnormalities on the zero-shot CT-CLIP embedding used in our pipeline. Every abnormality satisfies tail$_{1/2}${} $>$ top2, confirming that the misalignment is not driven by only a few outliers.

\begin{table}[!htb]
\vspace{-10pt}
\caption{Per-abnormality projection test on CT-CLIP zero-shot. All 18 abnormalities show tail$_{1/2}${} $>$ top2.}
\label{tab:supp-perfinding}
\centering
\scriptsize
\begin{tabular}{@{}lccc@{}}
\toprule
Abnormality & top2 AUC & tail$_{1/2}${} AUC & $\Delta$ \\
\midrule
Medical material & 0.532 & 0.770 & +0.238 \\
Arterial wall calcification & 0.514 & 0.708 & +0.194 \\
Coronary artery wall calc. & 0.501 & 0.713 & +0.212 \\
Lymphadenopathy & 0.519 & 0.677 & +0.158 \\
Lung nodule & 0.545 & 0.651 & +0.106 \\
Pericardial effusion & 0.514 & 0.817 & +0.303 \\
Hiatal hernia & 0.525 & 0.748 & +0.223 \\
Mosaic attenuation & 0.540 & 0.808 & +0.268 \\
Peribronchial thickening & 0.571 & 0.755 & +0.184 \\
Consolidation & 0.531 & 0.726 & +0.195 \\
Bronchiectasis & 0.537 & 0.774 & +0.237 \\
Interlob.\ septal thickening & 0.534 & 0.808 & +0.274 \\
Cardiomegaly & 0.571 & 0.779 & +0.208 \\
Emphysema & 0.521 & 0.711 & +0.190 \\
Atelectasis & 0.536 & 0.709 & +0.173 \\
Lung opacity & 0.542 & 0.672 & +0.131 \\
Pulm.\ fibrotic seq. & 0.519 & 0.682 & +0.163 \\
Pleural effusion & 0.535 & 0.787 & +0.253 \\
\bottomrule
\end{tabular}
\vspace{-10pt}
\end{table}

\section{Implementation Details}
\label{sec:supp-setup}

\subsection{Dataset and report preprocessing}
The released CT-RATE splits contain 47,149 training and 3,039 validation CT volumes. We exclude 742 training and 52 validation volumes without pre-computed organ embeddings, leaving 46,407 training and 2,987 validation volumes in our experiments. Each paired report contains Findings and Impression sections, which we concatenate into a single target sequence after basic text normalisation.

For retrieval database construction, organ-level report paragraphs (produced by a rule-based organ-section parser) are sentence-split by a regex tokeniser that segments on full stops and semicolons followed by whitespace. Fragments consisting solely of punctuation or bare organ-name headers are discarded. Sentences that do not map cleanly to the four target organs are assigned to an \emph{other} category and excluded from the retrieval indices. The final per-organ sentence database is built from training reports only, and \cref{tab:supp-db-stats} summarises its composition. Reassembling the organ-split paragraphs yields BLEU-1\,$=$\,0.992 against the original report, and per-organ Clinical F1 (Parser F1 in \cref{tab:supp-db-stats}) exceeds 0.95 for all four target organs, confirming faithful preservation of abnormality content.

\begin{table}[!htb]
\vspace{-10pt}
\caption{Per-organ sentence database statistics (training set only).}
\label{tab:supp-db-stats}
\centering
\scriptsize
\begin{tabular}{@{}lccccc@{}}
\toprule
Organ & Sentences & Unique reports & Avg.\ sents/report & Avg.\ words/sent & Parser F1 \\
\midrule
Lung      & 398{,}538 & 46{,}380 & 8.59 & 13.3 & 0.974 \\
Heart     &  82{,}025 & 42{,}777 & 1.92 &  8.4 & 0.957 \\
Esophagus &  50{,}096 & 39{,}977 & 1.25 & 11.8 & 0.990 \\
Aorta     &  41{,}632 & 25{,}093 & 1.66 & 10.0 & 0.955 \\
\midrule
Total     & 572{,}291 & ---      & ---  & --- & --- \\
\bottomrule
\end{tabular}
\vspace{-10pt}
\end{table}

\subsection{Training, evaluation, and checkpoint selection}
Clinical efficacy follows the CT-RATE protocol~\citep{Hamamci_2026}: a RadBERT classifier extracts 18 binary abnormality labels per report, and we report support-weighted Precision, Recall, and F1. BLEU, ROUGE-L, and METEOR are computed after lowercasing. LLaMA Score uses Llama-3.1-70B-Instruct as judge. Report-generation bootstrap 95\% CIs use 1{,}000 iterations, and the PCA analysis uses 200 resamples.
Both the 8B and 70B models follow the same two-stage training recipe: (1) train organ-specific projectors with the LLM frozen, then (2) LoRA fine-tune the language model after projector training. The base visual input comprises one global CT-CLIP embedding and four organ-specific ViSD-Boost embeddings, projected to the LLM hidden dimension by independent two-layer MLPs. All reported validation results correspond to the manually selected checkpoint with the best Clin-F1 on the validation set. To isolate whether the bottleneck lies in the generator or in the visual representation, the 8B and 70B variants are trained with the same visual interface, the same projector design, and the same evaluation protocol; both models receive the same five visual tokens, differing only in LLM scale.

\begin{table}[!htb]
\vspace{-10pt}
\caption{Per-abnormality Clinical F1 breakdown: base model (8B, no RAG) vs.\ AdaRAG-CT (8B, two-stage). $\Delta$: change due to adaptive retrieval augmentation.}
\label{tab:supp-perfinding-f1}
\centering
\scriptsize
\begin{tabular}{@{}lccc@{}}
\toprule
Abnormality & Base F1 & AdaRAG F1 & $\Delta$ \\
\midrule
\multicolumn{4}{@{}l}{\textit{Lung (11)}} \\
Lung nodule & 0.565 & 0.591 & +0.025 \\
Mosaic attenuation & 0.349 & 0.379 & +0.030 \\
Peribronch.\ thick. & 0.245 & 0.160 & $-$0.085 \\
Consolidation & 0.524 & 0.534 & +0.009 \\
Bronchiectasis & 0.096 & 0.151 & +0.054 \\
Interlob.\ septal thick. & 0.251 & 0.343 & +0.092 \\
Emphysema & 0.433 & 0.410 & $-$0.023 \\
Atelectasis & 0.437 & 0.436 & $-$0.001 \\
Lung opacity & 0.651 & 0.677 & +0.027 \\
Pulm.\ fibrotic seq. & 0.275 & 0.475 & \textbf{+0.200} \\
Pleural effusion & 0.641 & 0.748 & \textbf{+0.107} \\
\midrule
\multicolumn{4}{@{}l}{\textit{Heart (2)}} \\
Cardiomegaly & 0.440 & 0.452 & +0.012 \\
Pericardial effusion & 0.112 & 0.132 & +0.020 \\
\midrule
\multicolumn{4}{@{}l}{\textit{Aorta (2)}} \\
Arterial wall calc. & 0.654 & 0.688 & +0.034 \\
Coronary artery calc. & 0.592 & 0.666 & +0.073 \\
\midrule
\multicolumn{4}{@{}l}{\textit{Esophagus (1)}} \\
Hiatal hernia & 0.261 & 0.358 & \textbf{+0.097} \\
\midrule
\multicolumn{4}{@{}l}{\textit{Other (2)}} \\
Lymphadenopathy & 0.329 & 0.207 & $-$0.122 \\
Medical material & 0.240 & 0.066 & $-$0.174 \\
\midrule
\textbf{Weighted avg (Clin-F1)} & \textbf{0.455} & \textbf{0.480} & \textbf{+0.025} \\
\bottomrule
\end{tabular}
\vspace{-10pt}
\end{table}

\subsection{RAG token supervision and oracle-mixed training}
A dedicated \texttt{[RAG]} token is appended to the tokenizer vocabulary and fine-tuned jointly with the generation objective. At each perplexity-selected target, up to three MMR-selected training sentences are wrapped in \texttt{<|ret\_start|>\,\ldots\,<|ret\_end|>} and inserted before the target sentence. Context tokens are masked from the language-modeling loss (labels set to $-100$).
The context mode is sampled once per training report. Both 8B and 70B use oracle/real/no-context probabilities of $0.7/0.3/0.0$ and allow at most four \texttt{[RAG]} insertions per report. Offline oracle and real context construction use MMR with $\lambda{=}0.6$ and $\lambda{=}0.7$, respectively. Candidates with zero abnormality-label overlap are downweighted by a factor of 0.3, and repeated organ/normality axes receive a 0.3 penalty. Inference uses label-free MMR. Both models use LoRA $r{=}32$, $\alpha{=}64$, dropout 0.05, lr$\,{=}10^{-5}$, frozen visual projectors, 10{,}000 training steps, and a cosine schedule with 200 warmup steps; the effective batch sizes are 16 and 32 for 8B and 70B, respectively. At inference, each \texttt{[RAG]} trigger retrieves three sentences, with at most eight retrievals per report and a five-token cooldown.

\section{Additional Results}
\label{sec:supp-reproduced}

\subsection{Per-abnormality clinical F1 breakdown}
\Cref{tab:supp-perfinding-f1} decomposes the aggregate Clinical F1 improvement into individual abnormalities. AdaRAG-CT improves F1 on 13 of 18 abnormalities. The largest gains occur for abnormalities that require fine-grained descriptive detail: pulmonary fibrotic sequela (+0.200), pleural effusion (+0.107), and hiatal hernia (+0.097). These are abnormalities where the base model's impoverished visual channel fails to encode sufficient detail, and the retrieved textual context supplies the missing semantics. The few regressions (medical material $-$0.174, lymphadenopathy $-$0.122) involve abnormalities whose retrieval candidates are sparse or dominated by negative templates.

\subsection{Adaptive trigger statistics}
On the CT-RATE validation set (2{,}987 reports), the 8B AdaRAG-CT model emits an average of 1.48 \texttt{[RAG]} triggers per report (median 1). 22\% of reports receive zero triggers, showing that retrieval is not applied uniformly, while 61\% receive 1--2 triggers and 17\% receive 3 or more.

\subsection{Retrieval sanity checks}
\Cref{tab:supp-retrieval-sanity} compares the main-table AdaRAG-CT result with corrupt-context controls evaluated on the full CT-RATE validation set. The corrupt rows replace retrieved context with random training sentences or wrong-organ abnormal sentences and are averaged over three seeds. Corrupt contexts reduce Clinical F1 while leaving BLEU/ROUGE nearly unchanged, indicating that surface fluency does not imply clinically useful retrieval.

\begin{table}[!htb]
\vspace{-10pt}
\caption{Retrieval sanity checks on CT-RATE validation. Corrupt-context rows are averaged over seeds 42, 66, and 2026.}
\label{tab:supp-retrieval-sanity}
\centering
\scriptsize
\setlength{\tabcolsep}{4pt}
\begin{tabular}{@{}lcccccc@{}}
\toprule
Setting & Clin-P & Clin-R & Clin-F1 & B-1 & B-4 & R-L \\
\midrule
AdaRAG-CT 8B & 0.502 & 0.520 & 0.480 & 0.496 & 0.242 & 0.354 \\
Random context & 0.488 & 0.501 & 0.462 & 0.513 & 0.249 & 0.357 \\
Wrong-organ abnormal & 0.485 & 0.505 & 0.462 & 0.513 & 0.249 & 0.358 \\
\bottomrule
\end{tabular}
\vspace{-10pt}
\end{table}

\subsection{Reproduced baselines}
To ensure a fair comparison under consistent evaluation conditions, we reproduce CT-CHAT under our evaluation pipeline and report all metrics alongside our own models in \cref{tab:supp-reproduced}. The relative ranking is preserved under our protocol, while our base and adaptive models remain substantially stronger on clinical efficacy.

\begin{table}[!htb]
\vspace{-10pt}
\caption{Reproduced baselines under our unified evaluation protocol (CT-RATE validation set).}
\label{tab:supp-reproduced}
\centering
\scriptsize
\setlength{\tabcolsep}{3pt}
\begin{tabular}{@{}ll ccc cccc c @{}}
\toprule
& & \multicolumn{3}{c}{Clinical Efficacy} & \multicolumn{4}{c}{NLP} & \\
\cmidrule(lr){3-5} \cmidrule(lr){6-9}
Method & Params & F1 & P & R & B-1 & B-4 & R-L & MET & LLaMA \\
\midrule
CT-CHAT~\citep{Hamamci_2026} & 8B & 0.224 & 0.331 & 0.366 & 0.421 & 0.188 & 0.303 & 0.193 & 6.73 \\
CT-CHAT~\citep{Hamamci_2026} & 70B & 0.161 & 0.334 & 0.131 & 0.356 & 0.182 & 0.321 & 0.190 & 6.02 \\
\midrule
Ours (base) & 8B & 0.455 & 0.474 & 0.469 & 0.463 & 0.205 & 0.315 & 0.206 & 7.30 \\
AdaRAG-CT & 8B & 0.480 & 0.502 & 0.520 & 0.496 & 0.242 & 0.354 & 0.246 & 7.75 \\
Ours (base) & 70B & 0.405 & 0.468 & 0.373 & 0.449 & 0.213 & 0.334 & 0.208 & 7.10 \\
AdaRAG-CT & 70B & 0.426 & 0.483 & 0.413 & 0.497 & 0.250 & 0.361 & 0.232 & 7.53 \\
\bottomrule
\end{tabular}
\vspace{-10pt}
\end{table}

\subsection{Qualitative Analysis}
\label{sec:exp-qualitative}

\begin{figure}[!htb]
\vspace{-10pt}
  \centering
  \includegraphics[width=0.95\linewidth]{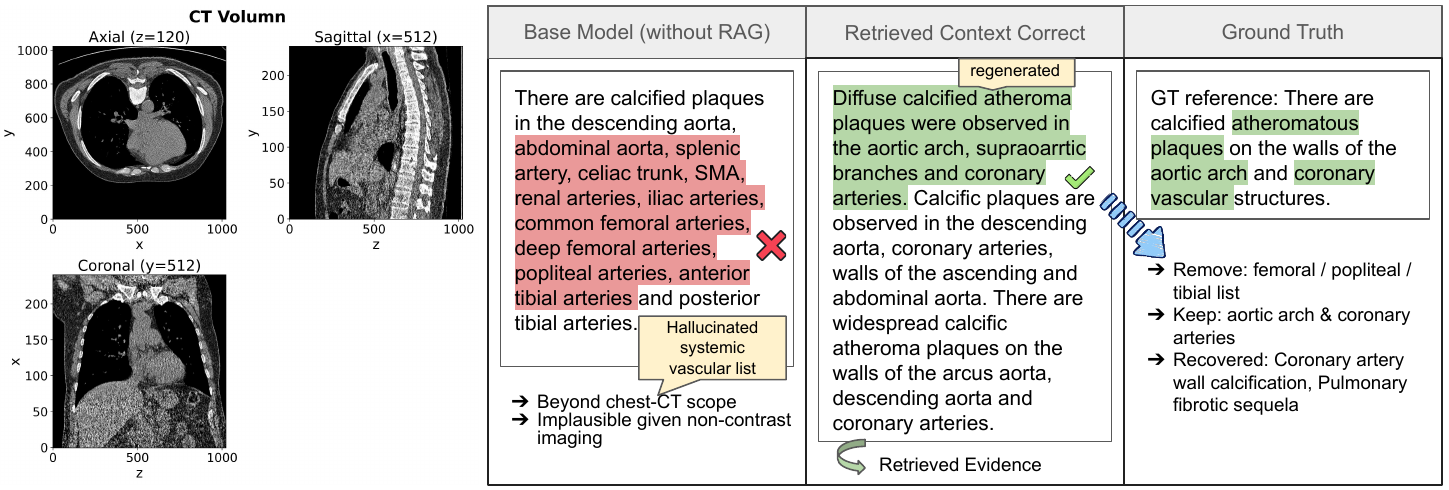}
  \caption{Qualitative comparison for an aorta abnormality. Left: base model output. Centre: AdaRAG-CT output with the regenerated sentence highlighted. Right: ground-truth report. Red highlights indicate hallucinated content. Green indicates correct clinical abnormalities.}
  \label{fig:qualitative}
\vspace{-10pt}
\end{figure}

\Cref{fig:qualitative} illustrates how the representational bottleneck manifests and how retrieval compensates. The base model, relying solely on the impoverished visual embedding, hallucinates a long list of peripheral vascular structures (femoral, popliteal, and tibial arteries) that are outside the chest-CT field of view and implausible under non-contrast imaging, a pattern consistent with the limited visual representation. When the adaptive \texttt{[RAG]} trigger fires, the retrieved evidence supplies case-relevant context centred on the aortic arch and coronary arteries. The model then regenerates the sentence, correctly focusing on aortic arch calcification and coronary vascular involvement, which aligns with the ground truth. This example illustrates that retrieval does not merely re-rank information already present in the visual input. It introduces clinically specific evidence that the visual embedding alone cannot provide.

\end{document}